\documentclass{article}
\pdfoutput=1
\usepackage{ijcai21}

\usepackage{times}
\usepackage{soul}
\usepackage{url}
\usepackage[hidelinks]{hyperref}
\usepackage[utf8]{inputenc}
\usepackage[small]{caption}
\usepackage{graphicx}
\usepackage{amsmath}
\usepackage{amsthm}
\usepackage{booktabs}
\usepackage{algorithm}
\usepackage{algorithmic}

\usepackage{multirow} 
\usepackage{multicol} 
\usepackage{arydshln}
\usepackage{amsfonts}

\title{CAT: Cross-Attention Transformer for One-Shot Object Detection}

\author{
Paper ID 3881
\affiliations
Anonymous
}

\author{
Weidong Lin$^{1,2}$\and
Yuyan Deng$^{1,2}$\and
Yang Gao$^{1,2}$\and
Ning Wang$^{1,2}$\and
Jinghao Zhou$^{1}$\and
Lingqiao Liu$^{3}$\and
Lei Zhang$^{1,2}$\and
Peng Wang$^{1,2}$\footnote{Corresponding Author}

\affiliations
$^1$School of Computer Science, Northwestern Polytechnical University, China \\
$^2$National Engineering Lab for Integrated Aero-Space-Ground-Ocean \\Big Data Application Technology, China \\
$^3$The University of Adelaide, Australia\\
\emails
\{weidong.lin, dengyuyan, gy7, ningw\}@mail.nwpu.edu.cn \\
jensen.zhoujh@gmail.com, lingqiao.liu@adelaide.edu.au \\
\{nwpuzhanglei, peng.wang\}@nwpu.edu.cn\\
}

\usepackage{color}

\begin{document}

\maketitle

\begin{abstract}
Given a query patch from a novel class, one-shot object detection aims to detect all instances of that class in a target image through the semantic similarity comparison. However, due to the extremely limited guidance in the novel class as well as the unseen appearance difference between query and target instances, it is difficult to appropriately exploit their semantic similarity and generalize well. To mitigate this problem, we present a universal Cross-Attention Transformer (CAT) module for accurate and efficient semantic similarity comparison in one-shot object detection. The proposed CAT utilizes transformer mechanism to comprehensively capture bi-directional correspondence  between any paired pixels from the query and the target image, which empowers us to sufficiently exploit their semantic characteristics for accurate similarity comparison. In addition, the proposed CAT enables feature dimensionality compression for inference speedup without performance loss. Extensive experiments on COCO, VOC, and FSOD under one-shot settings demonstrate the effectiveness and efficiency of our method, \emph{e.g.}, it surpasses CoAE, a major baseline in this task by 1.0\% in AP on COCO and runs nearly 2.5 times faster. Code will be available in the future.
\end{abstract}

\begin{figure}[t!]
\begin{center}
\includegraphics[width=1.\linewidth]{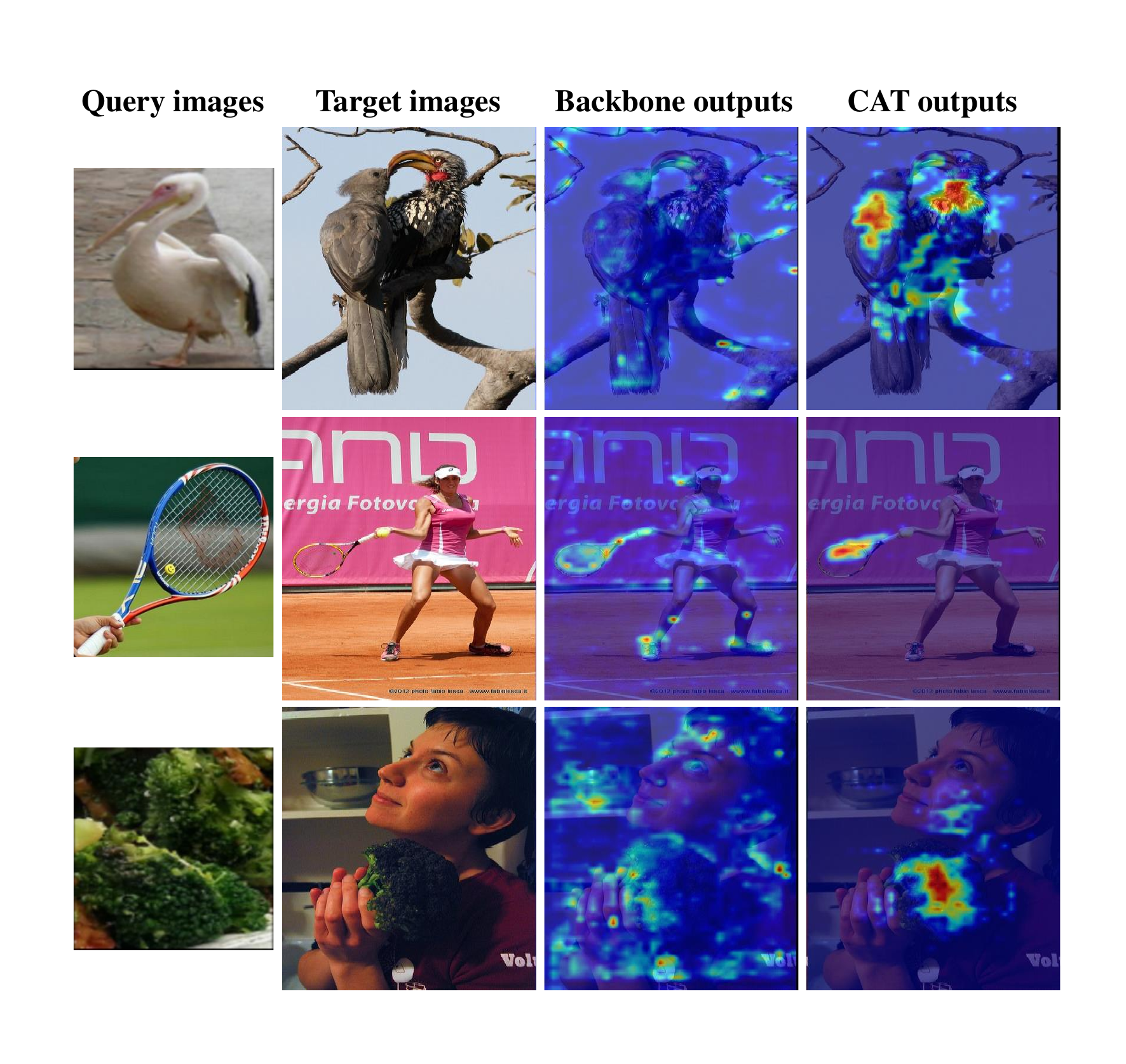}
\end{center}
  \caption{Visualization results of the intermediate feature maps. 
    We visualize the response maps of the input and output of our proposed CAT module in the last two columns.
    By capturing the bidirectional correspondence between query and target images,
    our CAT module significantly refines the response map and pays more attention on the objects with the same category of query objects.
  }
\label{fig:atten_map_pre}
\end{figure}

\section{Introduction} \label{intro}

Object detection is a fundamental task in computer vision domain, which 
aims to predict a bounding box with a category label for each instance of interest in an image. 
Although deep convolutional neural networks (DCNN) based object detection methods have achieved great success in recent years, 
their success heavily relies on a huge amount of annotated data, 
which is often difficult or even infeasible to collect in real applications due to the expensive annotation cost. 
Therefore, it is inevitable 
to cope with object detection for unseen classes with only a few annotated examples at test phase. 

In 
this study, we 
mainly focus on the most challenging problem, {\emph{i.e.}}, one-shot object detection. Given a novel class, there is only one query image with one annotated object, and a detector is then required to 
find all objects of the same category as the annotated object in a target image. 
Till now, some effective methods have been proposed, which mainly focus on building a two-stage paradigm~\cite{ren2016faster}. Specifically, in the first stage, the feature of query image and the target image is aggregated to exploit their semantic correspondence utilizing channel attention~\cite{hsieh2019one} or correlation filtering~\cite{fan2020fsod}. Then, a region proposal network is utilized to detect all candidate objects and real ones are ultimately located by a followed semantic similarity comparison based classifier. However, due to extremely limited guidance for the novel class (\emph{i.e.}, only one annotated sample) as well as the unseen appearance difference between the query object and the target one (\emph{e.g.}, that is often caused by the intra-class variation and different imaging endearments), these existing methods still fail to appropriately generalize well with pleasing performance.

To mitigate this problem, we revisit the one-shot object detection problem and attempt to explore the accurate semantic correspondence between the query object and the target image for performance enhancement. Considering that the great appearance difference often conceal their semantic correspondence into an unknown embedding space, we have to sufficiently exploit any detailed correspondence between two images. A direct way is to explore the relation between each sub-region from the query image and that in the target one. 
Following this idea, 
we propose a Cross-Attention Transformer (CAT) module and embed it into the two-stage detection paradigm for comprehensive exploration of the bidirectional correspondence between the target and query images.  
The proposed CAT module consists of two streams of interleaved transformers~\cite{vaswani2017attention}. Given the grid feature generated from a Siamese feature extractor, the two-stream transformer is utilized to exploit the bi-directional correspondence between any paired sub-regions from the query and the target image through computing the cross-attention between them. 
As shown in Figure~\ref{fig:atten_map_pre}, the CAT module can sufficiently exploit the semantic characteristics of each image as well as their grid-level correspondences, which will be beneficial for accurate similarity comparison in the second stage. In addition, due to sufficient information captured by the CAT module, the dimensionality of the final feature representation of each object can be effectively compressed without performance loss. To verify the effectiveness of the proposed method, we compare it with  state-of-the-art on three standard one-shot object detection benchmarks and observe significant performance and efficiency improvement.


In summary, this study mainly contributes in the following three aspects: 
\begin{itemize}
    \item We propose a CAT module which is able to sufficiently exploit the grid-level correspondence between the query and target image for accurate and efficient one-shot object detection. It is noticeable that the CAT module is an universal module which can be seamlessly plugged into other existing one-shot object detection frameworks.
    \item With the CAT module, we develop an effective one-shot detection network, which demonstrates state-of-the-art performance on three standard benchmarks for one-shot object detection. 
    \item By compressing the feature channels, the proposed model is capable of running  
    nearly $2.5$ times faster than the current state-of-the-art baseline CoAE~\cite{hsieh2019one} without performance degradation.

    
\end{itemize}

\begin{figure*}[t!]
\begin{center}
\includegraphics[width=1.\linewidth]{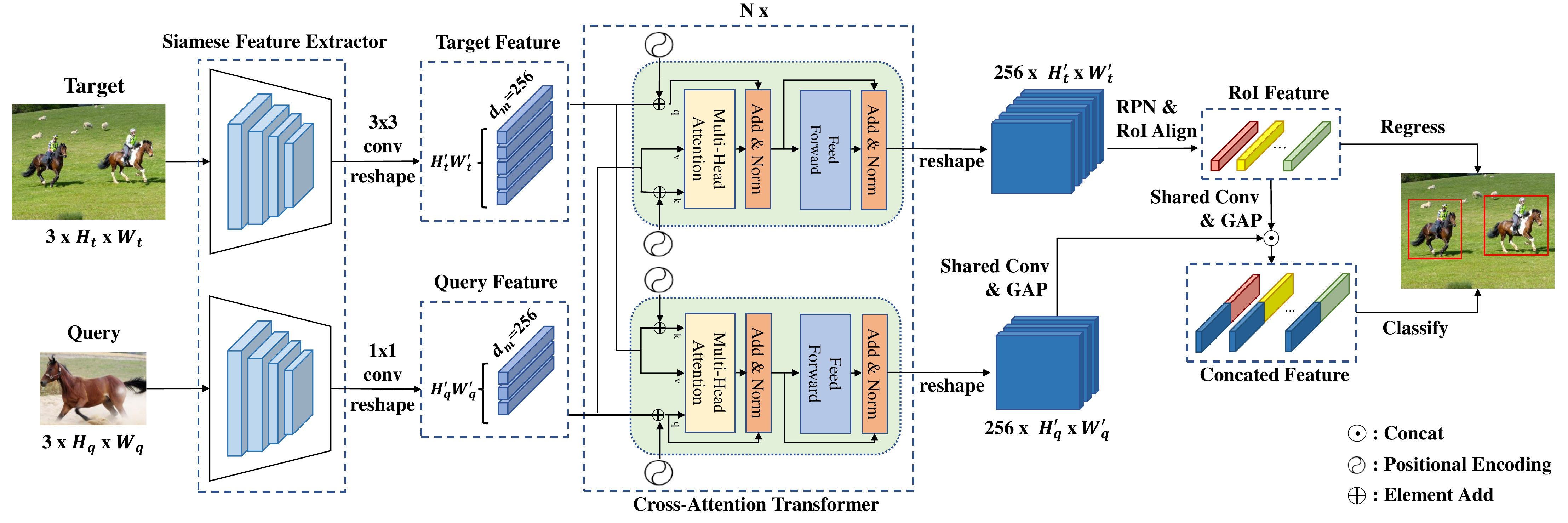}
\end{center}
\caption{The overall architecture of the proposed method for one-shot object detection. Our detector is composed of three parts. The first part is a shared ResNet-$50$ \protect\cite{he2016deep} backbone used to extract features of both the target and query images.  And the following part is our Cross-Attention Transformer (CAT) module that fuses the features from backbone and enhances the features of the regions which may be the same category as query in the target image, while the last part is the detection head with a regular RPN head and a R-CNN head like Faster R-CNN.}
\label{fig:overall architecture}
\end{figure*}

\section{Related Work}
In this section, we will briefly review two lines of research related to this study.


\subsection{Few-Shot Object Detection}
The 
key for few-shot object detection is to 
establish a similarity metric that can be 
appropriately generalize to 
unseen classes 
with a 
few labeled examples (\emph{i.e.} query set). 
Efforts have been made recently from different perspectives, such as transfer learning, metric learning and attention-based methods. 

Specifically, for transfer learning, \cite{chen2018lstd} presents the regularization techniques to 
relieve the over-fitting caused by 
directly transferring knowledge from a 
large auxiliary dataset to 
the novel classes. 
Another work ~\cite{kang2019few} develops a single-stage detector combined with a
meta-model that re-weights the importance of features from
the base model. 
For metric learning, ~\cite{karlinsky2019repmet} 
introduces a distance metric based 
classifier into the RoI module in the detector, which maps the objects 
into the universal embedding space. 
The attention-based methods focus on modelling the correspondence between target and query. ~\cite{hsieh2019one} designs a co-attention based model called CoAE which 
leverages the correlated features from the target and query for better generalization performance. ~\cite{fan2020fsod} introduces depth-wise convolution to get the attention feature map in the RPN phase and proposes the multi-relation detector to model different relationships in the R-CNN phase. ~\cite{osokin2020os2d} firstly performs dense correlation matching based on 
local features and then conducts spatial alignment 
and bi-linear resampling to compute the detection score. 

Our work lies on the third line of research, the attention-based methods. Different from previous work, our proposed CAT module empowers us to deeply exploit the grid-level bidirectional correspondence between target and query, using stacks of cross-attention transformer layers.

\subsection{Visual Transformer}
Witnessing that transformer becomes the de-facto standard in Natural Language Processing (NLP), recent literature commences introducing transformer-like networks into various computer vision tasks, 
including image recognition~\cite{dosovitskiy2020image,touvron2020training}, object detection~\cite{carion2020end,zhu2020deformable}, segmentation~\cite{ye2019cross}, visual question answering (VQA)~\cite{tan2019lxmert,su2019vl}. 
The Vision Transformer (ViT)~\cite{dosovitskiy2020image} directly feeds image patches into a transformer for image classification, which removes the need of any convolution operation. 
\cite{carion2020end} proposes DETR, a transformer encoder-decoder architecture that performs end-to-end object detection as set prediction. It does not rely on many manual components that required by traditional detectors, such as non-maximum suppression and anchor selection.
\cite{ye2019cross} proposes a cross-modal self-attention model to capture the long-range dependencies between language and visual features.
LXMERT~\cite{tan2019lxmert} and VL-BERT~\cite{su2019vl} are transformer-like visual-linguistic pretraining models that achieves superior performance on several vision-language tasks. 
To our best knowledge, our proposed model is the first attempt to employing transformers for the task of one-shot object detection. 
Moreover, it relies on a two-stream cross-attention architecture, rather than the commonly adopted self-attention mechanisms.


\section{Our Approach}


We formulate the one-shot object detection task as in \cite{hsieh2019one}.
Given a query image patch $p$ with its class label, the one-shot detector aims to detect all object instances of the same class in a target image $I$, where we assume that at least one instance exists in the target image.
We denote the set of classes in the testing phase (unseen classes) as $C_0$ while those in the training phase (seen classes) is $C_1$, 
and $C_0 \cap C_1= \emptyset$.
The model is trained with the annotated data of the seen classes, and generalize to unseen classes with a single query image. 


\subsection{Overall Architecture}
As shown in Figure~\ref{fig:overall architecture}, our proposed architecture is composed of three parts, including the feature extractor (backbone), the cross-attention module and the similarity-based detection head. 
At first, we adopt the ResNet-$50$ to extract features from both query image $I_q\in \mathbb{R}^{3\times H_q\times W_q}$ and target image $I_t\in \mathbb{R}^{3\times H_t\times W_t}$. 
Note that the backbone parameters are shared between query and target images. 
What needs to be especially explained is that we only use the first three blocks of ResNet-$50$ to extract feature maps with high resolutions. 
For the ease of representations, we denote $\phi(I_t) \in \mathbb{R}^{C\times H_t'\times W_t'}$ and $\phi(I_q)\in \mathbb{R}^{C\times H_q'\times W_q'}$ as the feature maps of target and query images respectively, where $\phi$ represents the backbone, $C=1024$, $H_t'=\frac{H_t}{16}$, $W_t'=\frac{W_t}{16}$, $H_q'=\frac{H_q}{16}$ and $W_q'= \frac{W_q}{16}$. After that, we use a $3\times 3$ convolution and a $1\times 1$ convolution to compress the number of channels of $\phi(I_t), \phi(I_q)$ from $1024$ to $d_m=256$. Both features are flattened in the dimension of spatial and further deeply aggregated by the CAT module with cross-attention mechanism as defined in the following formula:
\begin{equation}
(F_t, F_q)=\mathrm{CAT}(\phi(I_t)', \phi(I_q)'), \label{cat}
\end{equation}
where $\phi(I_t)' \in \mathbb{R}^{d_m\times H_t' W_t'}, \phi(I_q)' \in \mathbb{R}^{d_m\times H_q' W_q'}$ are the input sequences, and $F_t \in \mathbb{R}^{d_m\times H_t'\times W_t'}, F_q \in \mathbb{R}^{d_m\times H_q'\times W_q'}$ are the output feature maps after cross-attention.

In the end, RPN-based head takes as input the aggregated target features and generates proposals for further classification and regression. 
The features of proposals $p_1, p_2, \cdots, p_n$ extracted from $F_t$ by ROI align are fed into a regressor to obtain refined bounding boxes.
\begin{equation}
bbox_i=\Phi_r(\psi(F_t, p_i)), \label{reg}
\end{equation}
where $\Phi_r$ represents the regressor and $\psi$ represents the operation of ROI align.
For similarity-based classification, we first apply global average pooling on the RoI features and the aggregated query feature $F_q$,
and then concatenate them as the input of classifier $\Phi_c$. 
The classfication results $P(bbox_i), i=1, 2, \dots, n$ can be formulated as: 
\begin{equation}
P(bbox_i)=\Phi_c(\mathrm{Concat}(\mathrm{GAP}(\psi(F_t, p_i)), \mathrm{GAP}(F_q))). \label{reg}
\end{equation}

\subsection{Cross-Attention Transformer Module}
The cross-attention transformer (CAT) model is the key component of our proposed framework. Based on the transformer architecture, it models the bidirectional correspondences between grids of target and query images and performs dual feature aggregation for both target and query. 

The basic building block of transformer is the `Scaled Dot-Product Attention' defined as follows:
\begin{equation}
\mathrm{Attention}(Q, K, V)=\mathrm{softmax}(\frac{Q K^T}{\sqrt{d_k}})V, \label{atten}
\end{equation}
where $Q, K, V$ represent queries, keys and values, respectively. $d_k$ is the dimension of keys.

As described in \cite{vaswani2017attention}, Multi-Head Attention mechanism is further employed to jointly attend to information from different representation subspaces:
\begin{small}
\begin{equation}
\label{MultiHead}
\begin{split}
&\mathrm{MultiHead}(Q, K, V)=\mathrm{Concat(head_1, \cdots, head_M)}W^O, \\
&\mathrm{head_i}=\mathrm{Attention}(QW_i^Q, KW_i^K, VW_i^V),
\end{split}
\end{equation}
\end{small}
where $W_i^Q \in \mathbb{R}^{d_m \times d'}$, $W_i^K \in {\mathbb R}^{d_m \times d'}$, $W_i^K \in {\mathbb R}^{d_m \times d'}$ are the matrices to compute the so-called query, key and value embeddings respectively, and $W^O \in {\mathbb R}^{Md' \times d_m}$ is the projection matrix. In our work, we set $d'=\frac{d_m}{M}$, $d_m=256$ and $M=8$. 
 
After the Multi-Head Attention operation, the output is sent into a Feed-forward Network (FFN) module composed of two linear transformation with ReLU activation, defined as:
\begin{equation} 
\mathrm{FFN}(x)=\mathrm{max}(0, xW_1+b_1)W_2+b_2, \label{FFN}
\end{equation}
where $W_1, W_2$ and $b_1, b_2$ are the weight matrices and basis vectors respectively.

Recently, Carion~\cite{carion2020end} proposed a transformer-like model (DETR) for general object detection and obtain competing performance.
Although we also employ transformer in this work, there are still significant differences between DETR and our model.
Firstly, the challenges faced by the two models are different.
As a general object detector, DETR focuses on the discrimination between
foreground and background, and accurate bounding box regression.
On the contrary, the difficulty of one shot detection is mainly on the similarity-based comparison, 
rather than proposal generation~\cite{zhang2011proposal}.
Through experiments, we found that in many cases, one-shot detection models can produce accurate bounding boxes of salient objects but fails to assign correct class label.
To resolve their individual challenges, DETR and our model choose different model architectures.
DETR is built upon self-attention that explores long-range dependencies
between pixels of a single input image.
In contrast, our model relies on a two-stream architecture which performs cross-attention (Query-to-Target and Target-to-Query) to exploit the similarity between sub-regions of query and target images.

To be more specific, 
$X_t\in \mathbb{R}^{N_t \times d_m}$ and $X_q\in \mathbb{R}^{N_q \times d_m}$ represent the input sequences that are the flattened feature maps of target and query images respectively, as shown in Figure \ref{fig:overall architecture}.
Note that $N_t=H_t'\times W_t'$ and $N_q=H_q'\times W_q'$ are the lengths of the sequences.
Following ~\cite{carion2020end}, we use $sine$ function to generate spatial position encoding for input sequences $X_t$ and $X_q$.
In one stream of CAT, we let $Q = X_t$ and $K=V=X_q$ in equation \eqref{MultiHead}, and obtain the aggregated target feature.
This procedure can be summarized as:
\begin{small}
\begin{equation}
Y_t = \mathrm{Norm}(\widetilde{X_t}+FFN(\widetilde{X_t}))
\end{equation}
\end{small}
\begin{small}
\begin{equation}
\begin{split}
    \widetilde{X_t} = \mathrm{Norm}(X_t + P_t + \mathrm{MultiHead}(X_t+P_t, \\
    X_q+P_q, X_q))
\end{split}
\end{equation}
\end{small}where $P_t\in \mathbb{R}^{N_t \times d_m}, P_q\in \mathbb{R}^{N_q \times d_m}$ are the spatial position encodings corresponding to $X_t$ and $X_q$, respectively. 
In another stream, we set $Q=X_q$ and $K=V=X_t$ and generate 
$Y_q$, the aggregated query feature.
The above whole computation can be viewed as one layer of our proposed Cross-Attention Transformer, and the outputs of one layer will be the inputs of the next layer. 
In our work, we set the number of layers $N=4$. 

The outputs of CAT module are then reshaped to new feature maps $F_t$ and $F_q$ that share the same sizes as the origin feature maps, where $F_t$ is fed into the subsequent RPN and $F_q$ is used in similarity-based classification. 

\begin{table}[b!]
\small
    \centering
    \begin{tabular}{l|c|cc|cc}
        \hline
        \multirow{2}{*}{Method} &
        \multirow{2}{*}{FPS} &
        \multicolumn{2}{c|}{Unseen} &
        \multicolumn{2}{c}{Seen} \\
        \cline{3-6}
         & & AP & AP$50$ & AP & AP$50$ \\
        \hline
        CAT (One stream) & $18.4$ & $15.2$ & $25.3$ & $30.3$ & $48.6$ \\
        CAT (Two stream) & $16.3$ & ${\bf 16.5}$ & ${\bf 27.1}$ & ${\bf 31.3}$ & ${\bf 50.5}$ \\
        \hline
    \end{tabular}
    \caption{Ablation study of CAT on the COCO split $1$. 
    `Two stream' represents our model that performs both query-to-target and target-to-query attentions, while `One stream' represents a model that executes the query-to-target side.
    \label{table:results_coco_ablation_study}}
\end{table}

\begin{table}[b!]
\small
    \centering
    \begin{tabular}{l|cc|cc}
        \hline
        \multirow{2}{*}{Layers} &
        \multicolumn{2}{c|}{Unseen} &
        \multicolumn{2}{c}{Seen} \\
        \cline{2-5}
         & AP & AP$50$ & AP & AP$50$ \\
        \hline
        CAT ($3$ layers) & $15.8$ & $25.9$ & $31.2$ & $50.0$ \\
        CAT ($4$ layers) & ${\bf 16.5}$ & ${\bf 27.1}$ & $31.3$ & $50.5$ \\
        CAT ($5$ layers) & $16.5$ & $27.1$ & ${\bf 32.1}$ & ${\bf 51.6}$ \\
        CAT ($6$ layers) & $16.3$ & $27.1$ & $31.8$ & $51.3$ \\
        \hline
    \end{tabular}
    \caption{Results of Cross-Attention Transformer with different layers on COCO split 1.
    \label{table:results_layers}}
\end{table}

\begin{table}[b!]
\small
    \centering
    \begin{tabular}{l|cc|cc|cc}
        \hline
        \multirow{2}{*}{$d_m$} &
        \multirow{2}{*}{Params(M)} &
        \multirow{2}{*}{FPS} &
        \multicolumn{2}{c|}{Unseen} &
        \multicolumn{2}{c}{Seen} \\
        \cline{4-7}
         & & & AP & AP$50$ & AP & AP$50$ \\
        \hline
        $128$ & $12.74$ & $20.8$ & $14.6$ & $26.0$ & $29.0$ & $49.6$ \\
        $256$ & $19.10$ & $16.3$ & $16.5$ & $27.1$ & $31.3$ & $50.5$ \\
        $512$ & $37.67$ & $9.5$ & ${\bf 16.6}$ & ${\bf 27.3}$ & $32.1$ & $51.6$ \\
        $1024$ & $110.53$ & $4.9$ & $15.7$ & $25.8$ & ${\bf 32.7}$ & ${\bf 52.4}$ \\
        \hline
    \end{tabular}
    \caption{Results of different dimension of feature embeddings on COCO split 1.
    \label{table:results_channels}}
\end{table}

\section{Experiments}

Our experiments are conducted on the MS-COCO~\cite{lin2014microsoft}, PASCAL VOC and the recently released FSOD~\cite{fan2020fsod} dataset.
In Section~\ref{ID}, we first introduce implementation details.
Then we carry out ablation study and comparison with SOTA in Sections~\ref{ablation} and \ref{comparison} respectively.

\begin{table*}[t!]
\small
    \centering
    \begin{tabular}{l|cc|cc|cc|cc|cc}
        \hline
        \multirow{2}{*}{Method} &
        \multicolumn{2}{c|}{Split1} &
        \multicolumn{2}{c|}{Split2} &
        \multicolumn{2}{c|}{Split3} &
        \multicolumn{2}{c|}{Split4} &
        \multicolumn{2}{c}{Average} \\
        \cline{2-11}
         & AP & AP$50$ & AP & AP$50$ & AP & AP$50$ & AP & AP$50$ & AP & AP$50$ \\
        \hline
        SiamMask & - & $15.3$ & - & $17.6$ & - & $17.4$ & - & $17.0$ & - & $16.8$ \\
        CoAE & $11.8$ & $23.2$ & $12.2$ & $23.7$ & $9.3$ & $20.3$ & $9.4$ & $20.4$ & $10.7$ & $21.9$ \\
        CoAE (Reimp) & $15.1$ & $25.7$ & $15.3$ & $25.4$ & $11.0$ & $21.0$ & $12.5$ & ${\bf 21.7}$ & $13.5$ & $23.5$ \\
        Ours & ${\bf 16.5}$ & ${\bf 27.1}$ & ${\bf 16.6}$ & ${\bf 26.6}$ & ${\bf  12.4}$ & ${\bf 22.5}$ & ${\bf 12.6}$ & $21.4$ & ${\bf 14.5}$ & ${\bf 24.4}$ \\
        \hline
    \end{tabular}
    \caption{Results on the COCO dataset of unseen classes, we set the results of CoAE as our baseline. For fair comparisons, we re-implement CoAE on our code framework and report its results, `Reimp' represents our re-implemented model. 
    \label{table:results_coco}}
\end{table*}

\begin{table*}[t!]
\small
    \centering
    \begin{tabular}{l|cc|cc|cc|cc|cc}
        \hline
        \multirow{2}{*}{Method} &
        \multicolumn{2}{c|}{Split1} &
        \multicolumn{2}{c|}{Split2} &
        \multicolumn{2}{c|}{Split3} &
        \multicolumn{2}{c|}{Split4} &
        \multicolumn{2}{c}{Average} \\
        \cline{2-11}
         & AP & AP$50$ & AP & AP$50$ & AP & AP$50$ & AP & AP$50$ & AP & AP$50$ \\
        \hline
        SiamMask & - & $38.9$ & - & $37.1$ & - & $37.8$ & - & $36.6$ & - & $37.6$ \\
        CoAE  & $22.4$ & $42.2$ & $21.3$ & $40.2$ & $21.6$ & $39.9$ & $22.0$ & $41.3$ & $21.8$ & $40.9$ \\
        CoAE (Reimp)     & $31.2$ & ${\bf 51.3}$ & $27.3$ & $45.3$ & $27.7$ & $45.0$ & $28.8$ & $47.3$ & $28.8$ & $47.2$ \\
        Ours & ${\bf 31.3}$ & $50.5$ & ${\bf 28.8}$ & ${\bf 46.1}$ & ${\bf 28.9}$ & ${\bf 45.3}$ & ${\bf 29.6}$ & ${\bf 47.5}$ & ${\bf 29.7}$ & ${\bf 47.3}$ \\
        \hline
    \end{tabular}
    \caption{Results on the COCO dataset of seen classes.
    \label{table:results_coco_seen}}
\end{table*}

\begin{table*}[t!]
    \centering
    \resizebox{\textwidth}{!}{
    \begin{tabular}{l|cccccccccccccccc|c|cccc|c}
        \hline
        \multirow{2}{*}{Method} &
        \multicolumn{17}{c|}{Seen class} &
        \multicolumn{5}{c}{Unseen class} \\
        \cline{2-23}
        & plant & sofa & tv & car & bottle & boat & chair & person & bus & train & horse & bike & dog & bird & mbike & table & mAP & cow & sheep & cat & aero & mAP \\
        \hline
        SiamFC & $3.2$ & $22.8$ & $5.0$ & $16.7$ & $0.5$ & $8.1$ & $1.2$ & $4.2$ & $22.2$ & $22.6$ & $35.4$ & $14.2$ & $25.8$ & $11.7$ & $19.7$ & $27.8$ & $15.1$ & $6.8$ & $2.28$ & $31.6$ & $12.4$ & $13.3$ \\
        SiamRPN & $1.9$ & $15.7$ & $4.5$ & $12.8$ & $1.0$ & $1.1$ & $6.1$ & $8.7$ & $7.9$ & $6.9$ & $17.4$ & $17.8$ & $20.5$ & $7.2$ & $18.5$ & $5.1$ & $9.6$ & $15.9$ & $15.7$ & $21.7$ & $3.5$ & $14.2$ \\
        CompNet & $28.4$ & $41.5$ & $65.0$ & $66.4$ & $37.1$ & $49.8$ & $16.2$ & $31.7$ & $69.7$ & $73.1$ & $75.6$ & $71.6$ & $61.4$ & $52.3$ & $63.4$ & $39.8$ & $52.7$ & $75.3$ & $60.0$ & $47.9$ & $25.3$ & $52.1$ \\
        CoAE & $30.0$ & $54.9$ & $64.1$ & $66.7$ & $40.1$ & $54.1$ & $14.7$ & $60.9$ & $77.5$ & $78.3$ & $77.9$ & $73.2$ & $80.5$ & $70.8$ & ${\bf 72.4}$ & ${\bf 46.2}$ & $60.1$ & $83.9$ & $67.1$ & $75.6$ & $46.2$ & $68.2$ \\
        CoAE(Reimp) & ${\bf 47.3}$ & $61.8$ & ${\bf 72.1}$ & $83.0$ & ${\bf 56.6}$ & $63.1$ & $40.4$ & ${\bf 80.3}$ & ${\bf 81.3}$ & $80.6$ & $79.6$ & $77.1$ & $83.2$ & $75.0$ & $69.4$ & $45.5$ & ${\bf 68.5}$ & $84.3$ & ${\bf 76.5}$ & $81.5$ & $54.6$ & $74.2$ \\
        Ours & $44.2$ & ${\bf 65.5}$ & $67.1$ & ${\bf 83.9}$ & $54.2$ & ${\bf 66.8}$ & ${\bf 45.6}$ & $79.5$ & $76.8$ & ${\bf 82.3}$ & ${\bf 81.4}$ & ${\bf 78.5}$ & ${\bf 84.0}$ & ${\bf 76.7}$ & $71.0$ & $33.9$ & $68.2$ & ${\bf 84.8}$ & $75.6$ & ${\bf 83.7}$ & ${\bf 57.8}$ & ${\bf 75.5}$ \\
        \hline
    \end{tabular}}
    \caption{Results on the VOC dataset, we compare our model with several previous works and our baseline model CoAE.
    \label{table:results_voc}}
\end{table*}

\begin{table}[t!]
\small
    \centering
    \begin{tabular}{l|c|c|c}
        \hline
        Method & AP & AP$50$ & AP$75$ \\
        \hline
        CoAE (Reimp)     & $40.3$ & $63.8$ & $41.7$ \\
        Ours & ${\bf 42.0}$ & ${\bf 64.0}$ & ${\bf 44.2}$ \\
        \hline
    \end{tabular}
    \caption{Results on the FSOD dataset (unseen classes).
    \label{table:results_fsod}}
\end{table}

\subsection{Implementation Details}
\label{ID}

\textbf{Training Details}. Our network is trained with stochastic gradient descent (SGD) over $4$ NVIDIA RTX-2080Ti GPUs for $10$ epochs with the initial learning rate being $0.01$ and a mini-batch of $16$ images. The learning rate is reduced by a factor of $10$ at epoch $5$ and $9$, respectively. Weight decay and momentum are set as $0.0001$ and $0.9$, respectively. 
As in \cite{hsieh2019one}, the backbone ResNet-50 model is pretrained on a reduced training set of ImageNet in which all the COCO classes are removed to ensure that our model does note `foresee' any unseen class.
The target images are resized to have their shorter side being $600$ and their longer side less or equal to $1000$, and the query image patches are resized to a fixed size $128$x$128$.
We built our model on mmdetection~\cite{chen2019mmdetection}, which is a general object detection framework based on PyTorch.
Based on spatial-wise and channel-wise co-Attention,
CoAE~\cite{hsieh2019one} achieves the best performance over existing approaches and serve as a major baseline in our paper.
For strictly fair comparison, we re-implemented the CoAE model on 
the same mmdetection framework, and achieves significantly better results than the original author-provided version on all the three evaluated datasets. 
The reason may be better training strategies in mmdetection, such as multiply data augmentations and optimized pipeline.
\paragraph{Inference Details.} The same evaluation strategy as ~\cite{hsieh2019one} is applied for fair comparison. Specifically, we firstly shuffle the query image patches of that class with a random seed of target image ID, then sample the first five query image patches, we run our evaluations on these patches and take the average of these results as the stable statistics for evaluation.

\begin{figure*}[t!]
\begin{center}
\includegraphics[width=0.80\linewidth]{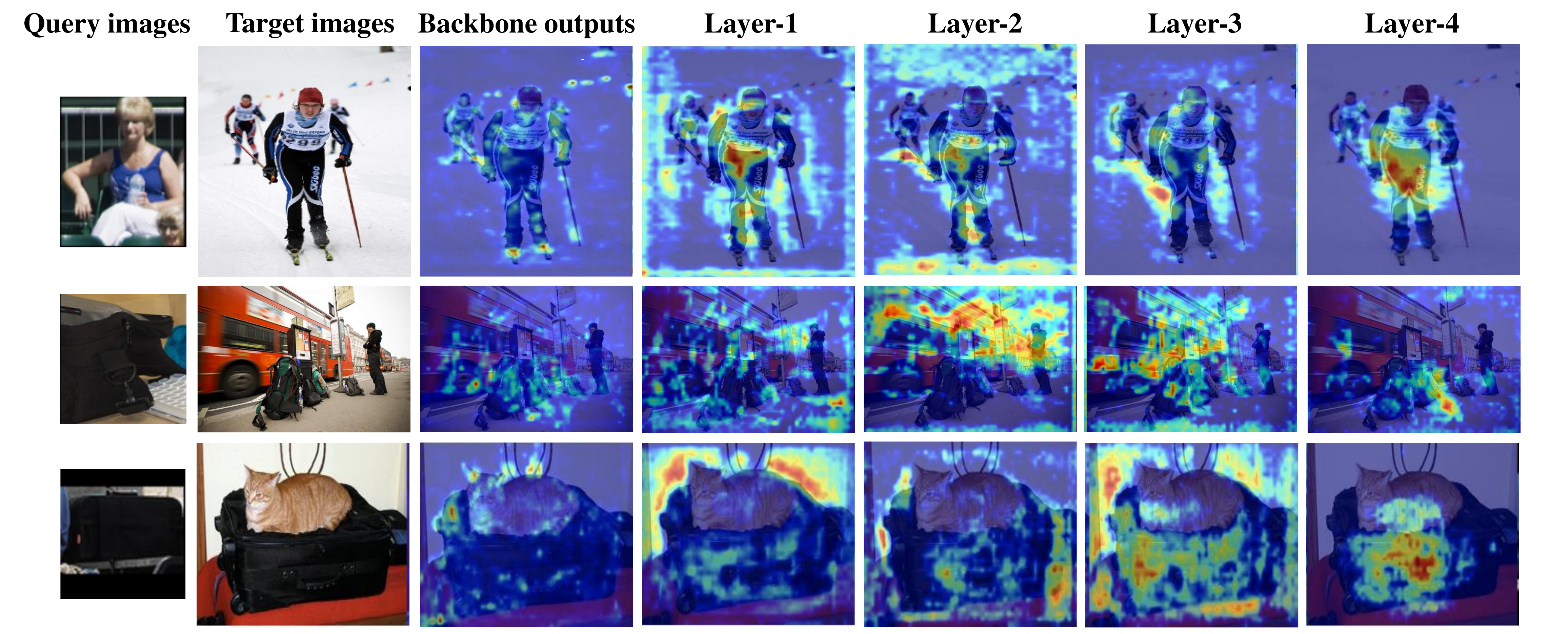}
\end{center}
  \caption{Visualization results of the intermediate feature maps. We visualize the outputs of each layer in our Cross-Attention Transformer on several target-query pairs.}
\label{fig:atten_layers}
\end{figure*}

\subsection{Ablation Study}
\label{ablation}
Since the Cross-Attention Transformer is the key component of our model, in this section we mainly explore the effect of this module with different hyper-parameters. For easy illustration, our ablation experiments are conducted on COCO split $1$ which will be discussed in Section~\ref{comparison}.

\paragraph{Transformer Structure.} Our CAT module consists of a stack of two-stream transformer layers, 
each of stream performing target-to-query or query-to-target attention and generating the corresponding target or query features. 
In Table~\ref{table:results_coco_ablation_study}, we compare the two-stream architecture with a one-stream transformer that only performs query-to-target attention and generates aggregated target features.
The results show that the the one-stream model incurs $1.3$ and $1.0$ AP drops on unseen and seen classes respectively, demonstrating the importance of bidirectional feature aggregation.

\paragraph{Number of CAT Layers.} We investigate the performance of CAT with different number of layers. As shown in Table~ \ref{table:results_layers}, we test the results of CAT with the number of layers ranging from $3$ to $6$.
The CAT with $4$ layers achieves the best performance on unseen classes, while on seen classes the best AP is obtained with the number of layers as $5$.
It can be found that increasing the number of layers does not always improve performance, which may be caused by the overfitting on seen classes.
Note that even using only $3$ layers, our model still outperforms the CoAE, demonstrating the superiority of the proposed method.
In the remaining experiments, we set the number of layers to $4$ by default. 

\paragraph{Dimension of Feature Embeddings.} 
Table \ref{table:results_channels} shows the results with different values of $d_m$ on COCO spit $1$. We also report their number of parameters and inference speed (FPS). 
From the results, we can find that reducing $d_m$ to $128$ will significantly decrease AP by $2$ points on unseen classes. 
The APs with $d_m = 256$ and $d_m = 512$ are close to each other, but
setting $d_m$ to $512$ will significantly increase the model size and slow down the inference speed. 
The results with $d_m = 1024$ shows an overfitting on seen classes.
To strike the balance between accuracy and speed, we set $d_m$ to $256$ in following experiments. 

\subsection{Comparison with State-of-the-Art}
\label{comparison}

\paragraph{MS-COCO.} Following the previous work~\cite{hsieh2019one}, we divide the $80$ classes of COCO dataset~\cite{lin2014microsoft} into four groups, alternately taking three groups as seen classes and one group as unseen classes. 
We use the `train $2017$' ($118$K images) split for training and minival (5K images) split for evaluation. 
We compare our method with SiamMask~\cite{michaelis2018one} and CoAE~\cite{hsieh2019one} in Tables~\ref{table:results_coco} and \ref{table:results_coco_seen}.
Besides the authors' release of CoAE model (denoted as CoAE in the Tables), we also re-implement this model in the mmdetection framework (denoted as CoAE(Reimp)) for strictly fair comparison. 
Note that CoAE (Reimp) is trained with the same strategies as our model and achieves better results than the original version, so it serves as a strong and fair baseline.  
Tables~\ref{table:results_coco} and \ref{table:results_coco_seen} show the comparison on unseen and seen classes, respectively.
Compared with the re-implemented CoAE model, our model achieves $1.0\%$ and $0.9\%$ improvements on the average AP and AP$50$ respectively.
As for seen classes, our model also achieves better performance that outperforms CoAE (Reimp) by $0.9$ AP point on average.

\paragraph{PASCAL VOC.} As for VOC~\cite{Everingham10}, we divide the $20$ classes into $16$ seen classes and $4$ unseen classes, where the choice of seen classes and unseen classes is consistent with ~\cite{hsieh2019one}. Note that our model is trained on the union set of VOC$2007$ train\&val sets and VOC$2012$ train\&val sets, while is evaluated on VOC$2007$. We evaluate the average precision of each category, and calculate mean average precision (mAP) of seen classes and unseen classes, respectively. Table~\ref{table:results_voc} shows the comparison with CoAE and other several baselines\cite{fu2020oscd,cen2018fully,8579033}, whose evaluation settings are consistent with ours. Our model outperforms
the re-implemented CoAE by $1.3$ mAP points on unseen classes and performs slightly worse ($0.3$ mAP) on seen classes, which presents a stronger generalization ability from seen classes to unseen classes.


\paragraph{FSOD.} The FSOD dataset~\cite{fan2020fsod} is specifically designed for few-shot object detection. It contains $1000$ categories, with $800$ for training and $200$ for test. We test the performance of our model and our re-implemented CoAE model on this dataset, with the same one-shot setting. Table~\ref{table:results_fsod} shows that our model outperforms CoAE by $1.7\%$ in AP and $2.5\%$ in AP\_{75} on novel classes.

\paragraph{Inference Speed.} 
Note that our model achieves superior accuracy with a much smaller 
dimension of features ($d_m = 256$) than that of the previous SOTA CoAE
($1024$).
On the other hand, the dot-product attention adopted by transformer is more parallelizable and space-efficient.
These two characteristics lead to a much faster inference speed:
on an NVIDIA RTX-2080Ti GPU, our model achieves $16.3$ FPS, while the speed of CoAE is only $5.9$ FPS that is nearly $2.5$ times slower than ours.




\subsection{Visualization of CAT layers}
For intuitively understanding our model, 
we visualize the intermediate feature maps according to the intensity of response. As shown in Figure~\ref{fig:atten_layers}, the first and second columns represent query and target images respectively, and the remaining columns correspond to the visualization of different CAT layers. 
Without incorporating any query information, the backbone outputs endow higher responses on salient objects or features. 
With the increase of layers and deeper aggregation of query information, the CAT outputs gradually focus on the objects of the same category as query. The visualization demonstrates the importance of our proposed CAT module on exploiting the correspondence between target and query.

\section{Conclusion} In this work, we propose a Cross-Attention Transformer module to deeply exploit bidirectional correspondence between the query and target pairs for one-shot object detection. 
By combining the proposed CAT module with a two-stage framework,
we construct a simple yet effective one-shot detector.
The proposed model achieves state-of-the-art performance on three one-shot detection benchmarks and meanwhile runs $2.5$ times faster than CoAE, a major strong baseline, demonstrating a superiority over both effectiveness and efficiency.

\appendix







\bibliographystyle{named}
\bibliography{ijcai21}

\begin{thebibliography}{}

\bibitem[\protect\citeauthoryear{{Carion} \bgroup \em et al.\egroup
  }{2020}]{carion2020end}
Nicolas {Carion}, Francisco {Massa}, Gabriel {Synnaeve}, Nicolas {Usunier},
  Alexander {Kirillov}, and Sergey {Zagoruyko}.
\newblock End-to-end object detection with transformers.
\newblock In {\em ECCV}, pages 213--229, 2020.

\bibitem[\protect\citeauthoryear{{Cen} and {Jung}}{2018}]{cen2018fully}
Miaobin {Cen} and Cheolkon {Jung}.
\newblock Fully convolutional siamese fusion networks for object tracking.
\newblock In {\em ICIP}, pages 3718--3722, 2018.

\bibitem[\protect\citeauthoryear{Chen \bgroup \em et al.\egroup
  }{2018}]{chen2018lstd}
Hao Chen, Yali Wang, Guoyou Wang, and Yu~Qiao.
\newblock Lstd: A low-shot transfer detector for object detection.
\newblock In {\em AAAI}, volume~32, 2018.

\bibitem[\protect\citeauthoryear{{Chen} \bgroup \em et al.\egroup
  }{2019}]{chen2019mmdetection}
Kai {Chen}, Jiaqi {Wang}, Jiangmiao {Pang}, Yuhang {Cao}, Yu~{Xiong}, Xiaoxiao
  {Li}, Shuyang {Sun}, Wansen {Feng}, Ziwei {Liu}, Jiarui {Xu}, Zheng {Zhang},
  Dazhi {Cheng}, Chenchen {Zhu}, Tianheng {Cheng}, Qijie {Zhao}, Buyu {Li}, Xin
  {Lu}, Rui {Zhu}, Yue {Wu}, Jifeng {Dai}, Jingdong {Wang}, Jianping {Shi},
  Wanli {Ouyang}, Chen~Change {Loy}, and Dahua {Lin}.
\newblock Mmdetection: Open mmlab detection toolbox and benchmark.
\newblock {\em arXiv preprint arXiv:1906.07155}, 2019.

\bibitem[\protect\citeauthoryear{Dosovitskiy \bgroup \em et al.\egroup
  }{2020}]{dosovitskiy2020image}
Alexey Dosovitskiy, Lucas Beyer, Alexander Kolesnikov, Dirk Weissenborn,
  Xiaohua Zhai, Thomas Unterthiner, Mostafa Dehghani, Matthias Minderer, Georg
  Heigold, Sylvain Gelly, et~al.
\newblock An image is worth 16x16 words: Transformers for image recognition at
  scale.
\newblock {\em arXiv preprint arXiv:2010.11929}, 2020.

\bibitem[\protect\citeauthoryear{Everingham \bgroup \em et al.\egroup
  }{2010}]{Everingham10}
M.~Everingham, L.~Van~Gool, C.~K.~I. Williams, J.~Winn, and A.~Zisserman.
\newblock The pascal visual object classes (voc) challenge.
\newblock {\em International Journal of Computer Vision}, 88(2):303--338, June
  2010.

\bibitem[\protect\citeauthoryear{Fan \bgroup \em et al.\egroup
  }{2020}]{fan2020fsod}
Qi~Fan, Wei Zhuo, Chi-Keung Tang, and Yu-Wing Tai.
\newblock Few-shot object detection with attention-rpn and multi-relation
  detector.
\newblock In {\em CVPR}, 2020.

\bibitem[\protect\citeauthoryear{Fu \bgroup \em et al.\egroup
  }{2020}]{fu2020oscd}
Kun Fu, Tengfei Zhang, Yue Zhang, and Xian Sun.
\newblock Oscd: A one-shot conditional object detection framework.
\newblock {\em Neurocomputing}, 2020.

\bibitem[\protect\citeauthoryear{He \bgroup \em et al.\egroup
  }{2016}]{he2016deep}
Kaiming He, Xiangyu Zhang, Shaoqing Ren, and Jian Sun.
\newblock Deep residual learning for image recognition.
\newblock In {\em CVPR}, pages 770--778, 2016.

\bibitem[\protect\citeauthoryear{{Hsieh} \bgroup \em et al.\egroup
  }{2019}]{hsieh2019one}
Ting-I {Hsieh}, Yi-Chen {Lo}, Hwann-Tzong {Chen}, and Tyng-Luh {Liu}.
\newblock One-shot object detection with co-attention and co-excitation.
\newblock In {\em NeurIPS}, volume~32, pages 2725--2734, 2019.

\bibitem[\protect\citeauthoryear{Kang \bgroup \em et al.\egroup
  }{2019}]{kang2019few}
Bingyi Kang, Zhuang Liu, Xin Wang, Fisher Yu, Jiashi Feng, and Trevor Darrell.
\newblock Few-shot object detection via feature reweighting.
\newblock In {\em CVPR}, pages 8420--8429, 2019.

\bibitem[\protect\citeauthoryear{Karlinsky \bgroup \em et al.\egroup
  }{2019}]{karlinsky2019repmet}
Leonid Karlinsky, Joseph Shtok, Sivan Harary, Eli Schwartz, Amit Aides, Rogerio
  Feris, Raja Giryes, and Alex~M Bronstein.
\newblock Repmet: Representative-based metric learning for classification and
  few-shot object detection.
\newblock In {\em CVPR}, pages 5197--5206, 2019.

\bibitem[\protect\citeauthoryear{{Li} \bgroup \em et al.\egroup
  }{2018}]{8579033}
B.~{Li}, J.~{Yan}, W.~{Wu}, Z.~{Zhu}, and X.~{Hu}.
\newblock High performance visual tracking with siamese region proposal
  network.
\newblock In {\em CVPR}, pages 8971--8980, 2018.

\bibitem[\protect\citeauthoryear{{Lin} \bgroup \em et al.\egroup
  }{2014}]{lin2014microsoft}
Tsung-Yi {Lin}, Michael {Maire}, Serge~J. {Belongie}, James {Hays}, Pietro
  {Perona}, Deva {Ramanan}, Piotr {Dollár}, and C.~Lawrence {Zitnick}.
\newblock Microsoft coco: Common objects in context.
\newblock In {\em ECCV}, pages 740--755, 2014.

\bibitem[\protect\citeauthoryear{{Michaelis} \bgroup \em et al.\egroup
  }{2018}]{michaelis2018one}
Claudio {Michaelis}, Ivan {Ustyuzhaninov}, Matthias {Bethge}, and Alexander~S.
  {Ecker}.
\newblock One-shot instance segmentation.
\newblock {\em arXiv preprint arXiv:1811.11507}, 2018.

\bibitem[\protect\citeauthoryear{Osokin \bgroup \em et al.\egroup
  }{2020}]{osokin2020os2d}
Anton Osokin, Denis Sumin, and Vasily Lomakin.
\newblock Os2d: One-stage one-shot object detection by matching anchor
  features.
\newblock {\em arXiv preprint arXiv:2003.06800}, 2020.

\bibitem[\protect\citeauthoryear{Ren \bgroup \em et al.\egroup
  }{2016}]{ren2016faster}
Shaoqing Ren, Kaiming He, Ross Girshick, and Jian Sun.
\newblock Faster r-cnn: Towards real-time object detection with region proposal
  networks.
\newblock {\em TPAMI}, 39(6):1137--1149, 2016.

\bibitem[\protect\citeauthoryear{Su \bgroup \em et al.\egroup
  }{2020}]{su2019vl}
Weijie Su, Xizhou Zhu, Yue Cao, Bin Li, Lewei Lu, Furu Wei, and Jifeng Dai.
\newblock {VL-BERT}: Pre-training of generic visual-linguistic representations.
\newblock In {\em ICLR}, 2020.

\bibitem[\protect\citeauthoryear{Tan and Bansal}{2019}]{tan2019lxmert}
Hao Tan and Mohit Bansal.
\newblock {LXMERT}: Learning cross-modality encoder representations from
  transformers.
\newblock In {\em EMNLP}, 2019.

\bibitem[\protect\citeauthoryear{Touvron \bgroup \em et al.\egroup
  }{2020}]{touvron2020training}
Hugo Touvron, Matthieu Cord, Matthijs Douze, Francisco Massa, Alexandre
  Sablayrolles, and Herv{\'e} J{\'e}gou.
\newblock Training data-efficient image transformers \& distillation through
  attention.
\newblock {\em arXiv preprint arXiv:2012.12877}, 2020.

\bibitem[\protect\citeauthoryear{{Vaswani} \bgroup \em et al.\egroup
  }{2017}]{vaswani2017attention}
Ashish {Vaswani}, Noam {Shazeer}, Niki {Parmar}, Jakob {Uszkoreit}, Llion
  {Jones}, Aidan~N. {Gomez}, Lukasz {Kaiser}, and Illia {Polosukhin}.
\newblock Attention is all you need.
\newblock In {\em NeurIPS}, volume~30, pages 5998--6008, 2017.

\bibitem[\protect\citeauthoryear{Ye \bgroup \em et al.\egroup
  }{2019}]{ye2019cross}
Linwei Ye, Mrigank Rochan, Zhi Liu, and Yang Wang.
\newblock Cross-modal self-attention network for referring image segmentation.
\newblock In {\em CVPR}, pages 10502--10511, 2019.

\bibitem[\protect\citeauthoryear{Zhang \bgroup \em et al.\egroup
  }{2011}]{zhang2011proposal}
Ziming Zhang, Jonathan Warrell, and Philip~HS Torr.
\newblock Proposal generation for object detection using cascaded ranking svms.
\newblock In {\em CVPR}, pages 1497--1504. IEEE, 2011.

\bibitem[\protect\citeauthoryear{Zhu \bgroup \em et al.\egroup
  }{2020}]{zhu2020deformable}
Xizhou Zhu, Weijie Su, Lewei Lu, Bin Li, Xiaogang Wang, and Jifeng Dai.
\newblock Deformable detr: Deformable transformers for end-to-end object
  detection.
\newblock {\em arXiv preprint arXiv:2010.04159}, 2020.

\end{thebibliography}

\end{document}